\documentclass[letterpaper]{article}
\usepackage{aaai}
\usepackage{times}
\usepackage{helvet}
\usepackage{courier}
\usepackage{algorithmic}
\usepackage{algorithm}
\usepackage{color}
\usepackage{graphicx}
\usepackage{multirow}

\def\iam{\tilde{\mathcal{A}}}
\def\l{\mathcal{L}}
\def\imp{{\tt ML-CBP}}
\newcommand{\ignore}[1]{{}}

\begin{document}
\title{Model-Lite Case-Based Planning}
\author{Hankz Hankui Zhuo$^a$, Subbarao Kambhampati$^b$, and Tuan Nguyen$^b$ \\
\mbox{$^a$Dept. of Computer Science, Sun Yat-sen University, Guangzhou, China} \\ 
\mbox{zhuohank@mail.sysu.edu.cn} \\
\mbox{$^b$Dept. of Computer Science and Engineering, Arizona State University, US} \\
\mbox{\{rao,natuan\}@asu.edu}
}

\maketitle
\begin{abstract}
There is increasing awareness in the planning community that
depending on complete models impedes the applicability of planning
technology in many real world domains where the burden of specifying
complete domain models is too high. In this paper, we consider a novel
solution for this challenge that combines generative planning on
incomplete domain models with a library of plan cases that are known to be
correct. While this was arguably the original motivation for case-based
planning, most existing case-based planners assume (and depend on)
from-scratch planners that work on complete domain models.
In contrast, our approach views the plan generated with respect to the
incomplete model as a ``skeletal plan'' and augments it with directed
mining of plan fragments from library cases. We will present the
details of our approach and present an empirical evaluation of
our method in comparison to a state-of-the-art case-based planner that
depends on complete domain models.
\end{abstract}

\section{Introduction}

Most work in planning assumes that complete domain models are
given as input in order to synthesize plans. However, there is
increasing awareness that  building
domain models at any level of completeness presents steep challenges for
domain creators. Indeed, recent work in web-service composition
(c.f. \cite{journal/aij/Bertoli10,conf/aaai/Hoffmann07}) and
work-flow management (c.f. \cite{journal/is/Blythe04}) suggest that
dependence on complete models can well be the real bottle-neck
inhibiting applications of current planning technology.

There has thus been interest in the so-called ``model-lite'' planning
approaches (c.f. \cite{conf/aaai/rao07}) that aim to synthesize plans even in the presence of
incomplete domain models. The premise here is that  while complete models cannot
be guaranteed, it is often possible for the domain experts to
put together reasonable but incomplete models. The challenge then is
to work with these incomplete domain models, and yet produce plans
that have a high chance of success with respect to the ``complete'' (but
unknown) domain model. This is only possible if the planner has access
to additional sources of knowledge besides the incomplete domain
model.

Interestingly, one of the original motivations for case-based planning
was also the realization that in many domains complete domain models
are not available. Over years however, case-based planning systems
deviated from this motivation and focused instead on ``plan reuse''
where the motivation is to improve the performance of a planner
operating with a complete domain model. In this paper, we return to the
original motivation by considering ``model-lite
case-based planning.''  In particular, we consider plan synthesis when
the planner has an incomplete domain theory, but has access to a
library of plans that ``worked'' in the past. This plan library can
thus be seen as providing additional knowledge of the domain over and
above the incomplete domain theory.

Our task is to effectively bring to bear this additional knowledge on
plan synthesis to improve the correctness of the plans generated. We
take a two stage process. First, we use the incomplete domain model to
synthesize a ``skeletal'' plan. Next, with the skeletal plan in hand,
we ``mine'' the case library for fragments of plans that can be
spliced into the skeletal plan to increase its correctness. The plan
improved this way is returned as the best-guess solution to the
original problem. We will describe the details of our framework,
called {\imp} and
present a systematic empirical evaluation of its effectiveness. We
compare the effectiveness of our model-lite case-based planner with
OAKPlan \cite{journals/aij/Serina10}, the current state-of-the-art model-complete case-based
planner.

We organize the paper as follows. We first review related work, and
then present the formal details of our framework. After that, we give a
detailed description of  {\imp} algorithm.  Finally, we
evaluate {\imp} in three planning domains, and compare its
performance to OAKPlan.

\section{Related Work}


As the title implies, our work is related both to case-based planning
and model-lite planning.
As mentioned in the introduction, our work is most similar to the
spirit of original case-based planning systems such as CHEF
\cite{book/Hammond89} and PLEXUS \cite{conf/aaai/alterman86}, which
viewed the case library as an extensional representation of the domain
knowledge. CHEF's use of case modification rules, for example, serves a similar
purpose as our use of incomplete domain models. Our work however
differs from CHEF in two ways. First, unlike us, CHEF assumes access
to a (more) complete domain model during its debugging stage. Second,
CHEF tries to adapt a specific case to the problem at hand, while our
work expands a skeletal plan with relevant plan fragments mined from
multiple library plans.
The post-CHEF case-based
planning work largely focused on having access to a from-scratch
planner operating on complete domain models
(c.f. \cite{priar,prodigy-analogy}). The most recent of this line of
work is OAKPlan \cite{journals/aij/Serina10}, which we compare
against.

The recent focus on planning with incomplete domain models originated
with the work on ``model-lite planning''
\cite{conf/aaai/rao07}. Approaches for model-lite planning must either
consider auxiliary knowledge sources or depend on long-term
learning. While our work views the case-library as the auxiliary
knowledge source, work by Nguyen et al. \cite{conf/icaps/Nguyen10} and
Weber et. al. \cite{cof/icaps/bryce11} assume that domain writers are
able to provide annotations about missing preconditions and
effects. It would be interesting to see if these techniques can be
combined with ours. One interesting question, for example, is whether
the case library can be compiled over time into such possible
precondition/effect annotations.
%
%

%

A third strand of research that is also related to our work is that of
action model learning. Work such as
\cite{journal/aij/Yang07,journal/aij/zhuo10,conf/aaai/Leslie05}
focuses on learning action models directly from observed (or
pre-specified) plan traces. The connection between this strand of work
and our work can be seen in terms of the familiar up-front
vs. demand-driven knowledge transfer: the learning methods attempt to
condense the case library directly into STRIPS models before using it
in planning, while we transfer knowledge from cases on a per-problem
basis.
Finally, in contrast, work such as \cite{conf/ijcai/Amir05},
as well as much of the reinforcement learning work \cite{rl-book}
focuses on learning models from trial-and-error
execution\footnote{This latter has to in general be limited to ergodic
  domains}. This too can be complementary to our work in that
execution failures can be viewed as opportunities to augment the
case-library (c.f. \cite{journal/jair/Ihrig97}).

%

\section{Problem Definition}
\begin{figure*}[!ht]
  \centering
  \includegraphics[width=0.68\textwidth]{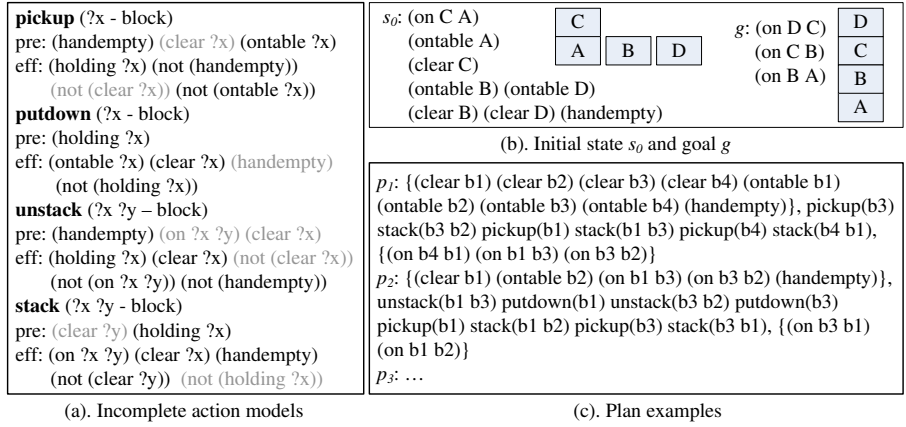}
  \caption{An input example of the {\imp} algorithm for the \emph{blocks} domain}\label{example:input}
\end{figure*}
A planning problem can be described as a triple $P=(\Sigma,s_0,g)$, where $s_0$ is an initial state, $g$ is a goal, and $\Sigma$ is defined by $\Sigma=(S, \mathcal A,\gamma)$, where $S$ is a set of states, $\mathcal A$ is a set of action models, and $\gamma$ is a transition function defined by $\gamma:S\times \mathcal A \rightarrow S$. A solution to a planning problem is an action sequence (or a plan) denoted by $(a_1,a_2,\ldots,a_n)$, where $a_i$ is an action. An \emph{action model} is defined as $(a,\textrm{PRE}(a), \textrm{ADD}(a), \textrm{DEL}(a))$, where $a$ is an action name with zero or more parameters, $\textrm{PRE}(a)$ is a precondition list specifying the condition under which $a$ can be applied, $\textrm{ADD}(a)$ is an adding list and $\textrm{DEL}(a)$ is a deleting list indicating the effects of $a$. Notice that we focus on the {\tt STRIPS} action model description \cite{journal/aij/Richard71} in this paper. An action model $a$ is called ``\emph{incomplete}'' when there are predicates missing in $\textrm{PRE}(a)$, $\textrm{ADD}(a)$, or $\textrm{DEL}(a)$. A set of incomplete action models is denoted by $\tilde{\mathcal{A}}$. An incomplete planning problem is denoted by $\tilde{P}=\langle s_0,g,\iam\rangle$. A plan example $p$ is composed by an initial state, a goal and an action sequence that transits the initial state and the goal, i.e., $p=\langle s_0,a_1,\ldots,a_n,g\rangle$, where $s_0$ is the initial state, $a_i$ is an action, and $g$ is the goal. We denote a set of plan examples by $E$.

Our planning problem in this paper is defined by: given as input a quadruple $\langle s_0, g, \tilde{\mathcal{A}}, E\rangle$, where $s_0$ is an initial state, and $g$ a goal, as described above, $\tilde{\mathcal{A}}$ is a set of incomplete action models, and $E$ is a plan example set, our {\imp} algorithm outputs a solution that transits $s_0$ and $g$.

An example input of our planning problem in \emph{blocks}\footnote{http://www.cs.toronto.edu/aips2000/} domain is shown in Figure \ref{example:input}, which is composed of three parts: incomplete action models (Figure \ref{example:input}(a)), an initial state $s_0$ and a goal $g$ (Figure \ref{example:input}(b)), and a plan example set (Figure \ref{example:input}(c)). In Figure \ref{example:input}(a), the dark parts indicate the missing predicates. In Figure \ref{example:input}(c), $p_1$ and $p_2$ are two plan examples, where initial states and goals are bracketed. An example output is a solution to the planning problem given in Figure \ref{example:input}, i.e., ``\emph{unstack(C A) putdown(C) pickup(B) stack(B A) pickup(C) stack(C B) pickup(D) stack(D C)}''.

\section{Our {\imp} Algorithm}
\begin{algorithm}[!ht]
\caption{Our {\imp} algorithm}\label{algorithm:main}
\textbf{Input:} $\tilde{P}=\langle s_0,g,\iam\rangle$, and a set of plan examples $E$. \\
\textbf{Output:} the plan $p^{sol}$ for solving the problem.

\begin{algorithmic}[1]
\STATE generate a set of causal pairs $\l$ with $\tilde{P}$;
\STATE build a set of plan fragments $\varphi$: \\ \quad\quad\quad\quad\quad $\varphi$=\emph{build\_fragments}($\l$, $E$);
\STATE mine a set of frequent plan fragments $\mathcal{F}$: \\ \quad\quad\quad\quad\quad $\mathcal{F}$=\emph{freq\_mining}($\varphi$);
\STATE $p^{sol}=\emptyset$;
\IF {\emph{concat\_frag}($p^{sol}$, $\l$, $\mathcal{F}$, $\tilde{P}$) = \TRUE}
  \RETURN $p^{sol}$;
\ELSE \RETURN NULL;
\ENDIF
\end{algorithmic}
\end{algorithm}
An overview of our {\imp} algorithm can be found in Algorithm \ref{algorithm:main}. We first generate a skeletal plan, presented by a set of causal pairs, based on $\langle s_0, g, \iam\rangle$. After that, we build a set of plan fragments based on plan examples and causal pairs, and then mine a set of frequent plan fragments with a specific threshold. These frequent fragments will be integrated together to form the final solution $p^{sol}$ based on causal pairs. Next, we describe each step in detail.

\subsection{Generate causal pairs}
Given the initial state $s_0$ and goal $g$, we generate a set of causal pairs $\l$. A causal pair is an action pair $\langle a_i,a_j\rangle$ that $a_i$ provides one or more conditions for $a_j$. The procedure to generate $\l$ is shown in Algorithm \ref{causal_links}.
\begin{algorithm}
\caption{Generate causal pairs}
\label{causal_links}
\textbf{input:} initial state $s_0$, goal $g$, incomplete action models $\iam$.
\textbf{output:} a set of causal pairs $\l$.
\begin{algorithmic}[1]
\STATE $\l=\emptyset$;
\FOR {each proposition $p\in g$}
  \STATE generate a plan, denoted by a set of causal pairs $\l'$, to transit $s_0$ to $p$;
  \STATE $\l=\l\cup\l'$;
\ENDFOR
\RETURN $\l$;
\end{algorithmic}
\end{algorithm}
Note that, in step 3 of Algorithm \ref{causal_links}, $\l'$ is an
empty set if $p$ cannot be achieved. In other words, skeletal plans
may not provide any guidance for some top level goals. Actions in causal pairs $\l$ is
viewed as a set of \emph{landmarks} for helping construct the final
solution, as will be seen in the coming sections. We show an example
of the generated causal pairs in Example 1. \\ \textbf{Example 1:}
\emph{As an example, causal pairs generated for the planning problem
  given in Figure \ref{example:input} is \{$\langle$pickup(B),stack(B
  A)$\rangle$, $\langle$ unstack(C A), stack(C B)$\rangle$, $\langle$
  pickup(D), stack(D C)$\rangle$\}.}

\subsection{Creating Plan Fragments}
In the procedure ``\emph{build\_fragments}'' of Algorithm \ref{algorithm:main}, we would like to build a set of plan fragments $\varphi$ by building mappings between ``objects'' in $\langle s_0,g\rangle$ of $\tilde{P}$ and $\langle s_0^i,g^i\rangle$ of a plan example $p_i\in E$. In other words, a mapping, denoted by $m$, is composed of a set of pairs $\{\langle o',o\rangle\}$, where $o'$ is an object (i.e., an instantiated parameter) from plan example $p_i$, and $o$ is an object from $\tilde{P}$. We can apply mapping $m$ to a plan example $p_i$, whose result is denoted by $p_i|_m$, such that $s_0^i|_m$ and $s_0$ share common propositions, likewise for $g^i$ and $g$. We measure a mapping $m$ by the number of propositions shared by initial states $s_0^i|_m$ and $s_0$, and goals $g^i$ and $g$. We denote the number of shared propositions by $\lambda(p_i,m)$, i.e., \[\lambda(p_i,m)=|(s_0^i|_m)\cap s_0|+|(g^i|_m)\cap g|.\] An example to demonstrate how to calculate $\lambda$ is given as follows.\\
\textbf{Example 2:} \emph{In Figure \ref{example:input}, a possible mapping $m$ between $\langle s_0,g\rangle$ and $\langle s_0^1, g^1\rangle$ of $p_1$ is $\{\langle b4,D\rangle,\langle b1,C\rangle,\langle b3,B\rangle,\langle b2,A\rangle\}$. The result of applying $m$ to $s_0^1$ is $s_0^1|_m$=\{(clear C)(clear A)(clear B)(clear D)(ontable C)(ontable A)(ontable B)(ontable D)(handempty)\}. Likewise, we can calculate the result of applying $m$ to $g^1$. It is not difficult to see that $\lambda(p_1,m)=|(s_0^1|_m)\cap s_0|+|(g^1|_m)\cap g|=10$.}

It is possible that there are many different mappings between $\tilde{P}$ and $p_i$. We choose a mapping $m^*$ with the largest $\lambda$ to maximally map $p_i$ to $\tilde{P}$, i.e., $m^*=\arg\max_m\lambda(p_i,m)$. We assume that all propositions are ``equally'' important in describing states. The more common propositions $\tilde{P}$ and $p_i$ share, the more ``similar'' they are. Note that mappings between objects of the same types are subject to the constraint that they should have the set of ``features'' in the domain, defined by unary predicates of the corresponding types. For instance, ``b3'' can be mapped to ``B'' in our running example since both of them are the two blocks having the same features ``on table'' and ``clear'' in the two problems. In practice, we find that this requirement significantly reduces the amount of mappings that need to be considered, actually allowing us to find $m^*$ in a reasonable running time.

We apply $m^*$ to $p_i$ to get a new plan example $p_i|_{m^*}$, which is denoted by $(a_1^i,a_2^i,\ldots,a_n^i)$. We scan the action sequence from $a_1$ to $a_n$ to get subsequences that satisfies the constraint that all the objects in the subsequences should be in $\tilde{P}$. We call these subsequences \emph{plan fragments}. We can build a set of plan fragments using plan examples $E$.\\
\textbf{Example 3:} \emph{In Example 2, we find that $m^*=\{\langle b4,D\rangle,\langle b1,C\rangle,\langle b3,B\rangle,\langle b2,A\rangle\}$. Thus, $p_1|_{m^*}$ is ``pickup(B) stack(B A) pickup(C) stack(C B) pickup(D) stack(D C)'', which is a plan fragment. For $p_2$ in Figure \ref{example:input}, $m^*$ is $\{\langle b3,C\rangle,\langle b1,B\rangle,\langle b2,A\rangle\}$. Thus, $p_2|_{m^*}$ is ``unstack(B C) putdown(B) unstack(C A) putdown(C) pickup(B) stack(B A) pickup(C) stack(C B)'', which is also a plan fragment.}

\subsection{Mining Frequent Plan Fragments}
In step 3 of Algorithm \ref{algorithm:main}, we aim at building a set of frequent plan fragments $\mathcal{F}$ using the procedure ``\emph{freq\_mining}''. Given that there will not be any function perfectly mapping the two planning problems, our intuition is that a plan fragment occurring multiple times in different plan examples increases our confidence on both the quality of the mapping between objects involved and the success of reusing the fragment as part of a solution plan for the problem being solved. We thus borrow the notion of frequent patterns defined in \cite{journal/ml/Zaki01,journal/tkde/pei04} to use for mining our frequent plan fragments. The problem of mining sequential patterns can be stated as follows. Let $\mathcal{I}=\{i_1,i_2,\ldots,i_n\}$ be a set of $n$ items. We call a subset $X\subseteq \mathcal{I}$ an itemset and $|X|$ the size of $X$. A sequence is an ordered list of itemsets, denoted by $s=\langle s_1,s_2,\ldots,s_m\rangle$, where $s_j$ is an itemset. The size of a sequence is the number of itemsets in the sequence, i.e., $|s|=m$. The length $l$ of a sequence $s=(s_1,s_2,\ldots,s_m)$ is defined as $l=\sum_{i=1}^m|s_i|$. A sequence $s_a=(a_1,a_2,\ldots,a_n)$ is a \emph{subsequence} of another sequence $s_b=(b_1,b_2,\ldots,b_m)$ if there exist integers $1\leq i_1<i_2<\ldots<i_n\leq m$ such that $a_1\subseteq b_{i_1},a_2\subseteq b_{i_2},\ldots,a_n\subseteq b_{i_n}$, denoted by $s_a\sqsubseteq s_b$. A sequence database $S$ is a set of tuples $\langle sid,s\rangle$, where $sid$ is a $sequence_id$ and $s$ is a sequence. A tuple $\langle sid,s\rangle$ is said to \emph{contain} a sequence $a$, if $a$ is a subsequence of $s$. The support of a sequence $a$ in a sequence database $S$ is the number of tuples in the database containing $a$, i.e., \[sup_S(a)=|\{\langle sid,s\rangle|(\langle sid,s\rangle\in S)\cap(a\sqsubseteq s)\}|.\] Given a positive integer $\delta$ as the support threshold, we call $a$ a \emph{frequent} sequence if $sup_S(a)\geq \delta$. Given a sequence database and the support threshold, frequent sequential pattern mining problem is to find the complete set of sequential patterns whose support is larger than the threshold.

We view each action of plan fragments as an itemset, and a plan fragment as a sequence, which suggests plan fragments can be viewed as a sequence database. Note that in our case an itemset has only one element, and the indices of those in the subsequence are continuous. We fix a threshold $\delta$ and use the SPADE algorithm \cite{journal/ml/Zaki01} to mine a set of frequent patterns. There are many frequent patterns which are subsequences of other frequent patterns. We eliminate these ``subsequences'' and keep the "maximal" patterns, i.e., those with the longest length, as the final set of frequent plan fragments $\mathcal{F}$. \\
\textbf{Example 4:} \emph{In Example 3, if we set $\delta$ to be 2 and 1, the results are shown below (frequent plan fragments are partitioned by commas):
\begin{center}
\begin{tabular}{|p{8cm}|}
\hline
\textbf{plan fragments:} \\
1. pickup(B) stack(B A) pickup(C) stack(C B) pickup(D) stack(D C) \\
2. unstack(C A) putdown(C) pickup(B) stack(B A) pickup(C) stack(C B) \\
\textbf{frequent plan fragments $\mathcal{F}$ ($\delta=2$):} \\
\{pickup(B) stack(B A) pickup(C) stack(C B)\} \\
\textbf{frequent plan fragments $\mathcal{F}$ ($\delta=1$):} \\
\{pickup(B) stack(B A) pickup(C) stack(C B) pickup(D) stack(D C), unstack(B C)putdown(B)unstack(C A) putdown(C) pickup(B) stack(B A) pickup(C) stack(C B)\} \\
\hline
\end{tabular}
\end{center}}
Note that the following frequent patterns are eliminated when $\delta=2$ (likewise when $\delta=1$): \{\emph{pickup(B), stack(B A), pickup(C), stack(C B), pickup(B) stack(B A), stack(B A) pickup(C), pickup(C) stack(C B), pickup(B) stack(B A) pickup(C), stack(B A) pickup(C) stack(C B)}\}.

\subsection{Generating Final Solution}
In steps 4-6 of Algorithm \ref{algorithm:main}, we generate the final solution using frequent plan fragments generated by step 3. We address the procedure \emph{concat\_frag} by Algorithm \ref{algorithm:frag}. In Algorithm \ref{algorithm:frag}, we scan each causal pair in $\l$ and each frequent plan fragment in $\mathcal{F}$; if a plan fragment contains an action (or both actions) of a causal pair, we append the plan fragment to the final solution $p^{sol}$ and remove all the causal pairs that are satisfied by the new $p^{sol}$; and then recursively call the procedure \emph{concat\_frag} until the solution is found, i.e., $\l = \emptyset$, or no solution is found, i.e., the procedure returns \emph{false} ($\l \neq \emptyset$).

\begin{figure}[!ht]
  \centering
  \includegraphics[width=0.34\textwidth]{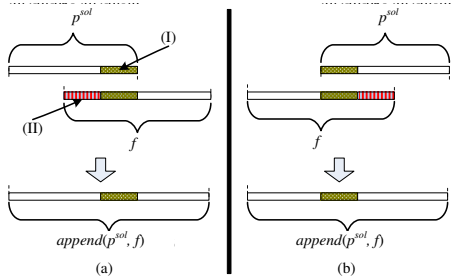}
  \caption{(a). $f$ is concatenated at the end of $p^{sol}$; (b). $f$ is concatenated at the beginning of $p^{sol}$; Part (I) is the maximal action subsequence; Part (II) is the action subsequence that is different from $p^{sol}$.}\label{figure:concat}
\end{figure}

\begin{algorithm}
\caption{\emph{concat\_frag}($p^{sol}$, $\l$, $\mathcal{F}$,$\tilde{P}$);}\label{algorithm:frag}
\textbf{input:} a plan $p^{sol}$, a set of causal pairs $\l$, a set of frequent plan fragments $\mathcal{F}$, and an incomplete problem; \\
\textbf{output:} \textbf{true} or \textbf{false}.

\begin{algorithmic}[1]
\IF {$\l = \emptyset$}
  \STATE $p^{sol}=remove\_first\_actions(p^{sol},\tilde{P})$;
  \STATE $p^{sol}=remove\_last\_actions(p^{sol},\tilde{P})$;
  \IF {$p^{sol}$ is executable based on $\tilde{P}$}
     \RETURN \TRUE;
  \ELSE
     \RETURN \FALSE;
  \ENDIF
\ENDIF
\FOR {each pair $\langle a_i,a_j\rangle\in\l$ and each $f\in\mathcal{F}$}
  \IF {($a_i\in f \vee a_j\in f$) and \emph{share}($p^{sol}$, $f$) =\TRUE}
    \STATE $p^{sol'}$=\emph{append}($p^{sol}$, $f$);
    \STATE $\l'$=\emph{removelinks}($p^{sol'}$, $\l$);
    \STATE $\mathcal{F}'\leftarrow\mathcal{F}-\{f\}$;
    \IF {\emph{concat\_frag}($p^{sol'}$, $\l'$, $\mathcal{F}$,$\tilde{P}$) =\TRUE}
      \RETURN \TRUE;
    \ENDIF
  \ENDIF
\ENDFOR
\RETURN \FALSE
\end{algorithmic}
\end{algorithm}
In step 2 of Algorithm \ref{algorithm:frag}, we repeatedly remove the first action of $p^{sol}$ that cannot be applied in $s_0$. In step 3 of Algorithm \ref{algorithm:frag}, we repeatedly remove the last action of $p^{sol}$ that deletes propositions of goal $g$. After steps 2 and 3, the remainder plan can be executed from $s_0$ to $g$ using $\iam$, then the algorithm returns \emph{true}, otherwise, returns \emph{false}. In step 11 of Algorithm \ref{algorithm:frag}, the procedure \emph{share} returns \emph{true} if $p^{sol}$ is empty or $p^{sol}$ and $f$ share a common action subsequence. That is to say, two plan fragments are concatenated only if they have some sort of \emph{connection}, which is indicated by common action subsequence. In step 12 of Algorithm \ref{algorithm:frag}, we concatenate $p^{sol}$ and $f$ based on their maximal common action subsequence, which is viewed as the strongest connection between them. Note that the common action subsequence should start from the beginning of $p^{sol}$ OR end at the end of $p^{sol}$. In other words, $f$ can be concatenated at the end of $p^{sol}$ or at the beginning, as is shown in Figure \ref{figure:concat}. In step 13 of Algorithm \ref{algorithm:frag}, the procedure \emph{removelinks} remove all causal pairs in $\l$ that are ``satisfied'' by $p^{sol}$. The result is denoted by $\l'$. Example 5 demonstrates how to generate final solutions.

\textbf{Example 5:} \emph{In Example 4, we have two frequent plan fragments by setting $\delta=1$. We concatenate these two fragments together. The result is shown as follows. The boldfaced part is the actions shared by fragments 1 an d 2. The concatenating result is shown in the third row. After concatenating, we can see that all the causal pairs in $\l$ is satisfied and will be removed according to step 13 of Algorithm \ref{algorithm:frag}. Furthermore, according to steps 2 and 3 of Algorithm \ref{algorithm:frag}, the first two actions are removed since they cannot be applied in $s_0$, and no action is removed at the end of the plan since no action deletes propositions of $g$. The result is shown in the fourth row. The result is executable from $s_0$ to $g$, which means it is the final solution.
\begin{tabular}{|p{8cm}|}
  \hline
  fragment 1: \textbf{pickup(B) stack(B A) pickup(C) stack(C B)} pickup(D) stack(D C) \\
  \hline
  fragment 2: unstack(B C) putdown(B) unstack(C A) putdown(C) \textbf{pickup(B) stack(B A) pickup(C) stack(C B)}\\
  \hline
  result: unstack(B C) putdown(B) unstack(C A) putdown(C) \textbf{pickup(B) stack(B A) pickup(C) stack(C B)} pickup(D) stack(D C) \\
  \hline
  solution: unstack(C A) putdown(C) \textbf{pickup(B) stack(B A) pickup(C) stack(C B)} pickup(D) stack(D C) \\
  \hline
\end{tabular}
}

\section{Experiments}
\subsection{Dataset and Criterion}
We evaluate our {\imp} algorithm in three planning domains: \emph{blocks}$^2$, \emph{driverlog}\footnote{http://planning.cis.strath.ac.uk/competition/} and \emph{depots}$^3$. In each domain, we generate from 40 to 200 plan examples using a classical planner such as FF planner\footnote{http://members.deri.at/$\sim$joergh/ff.html} and solve 100 new planning problems based on different percentages of completeness of domain models. For example, we use $\frac{4}{5}$ to indicate one predicate is missing among five predicates of the domain.

We define the accuracy of our {\imp} algorithm as the percentage of correctly solved planning problems. Specifically, we exploit {\imp} to generate a solution to a planning problem, and execute the solution from the initial state to the goal. If the solution can be successfully executed starting from the initial state, and the goal is achieved, then the number of correctly solved problems is increased by one. The accuracy, denoted by $\lambda$, can be computed by $\lambda=\frac{N_c}{N_t}$, where $N_c$ is the number of correctly solved problems, and $N_t$ is the number of total testing problems. Note that when testing the accuracy of {\imp}, we assume that we have \emph{complete} domain models available for executing generated solutions. It is easy to see that the larger the accuracy $\lambda$ is, the better our {\imp} algorithm functions.

\subsection{Experimental Results}
We would like to evaluate {\imp} in the following aspects: (1) the change of accuracies with respect to different number of plan examples; (2) the change of accuracies with respect to different percentages of completeness; (3) the change of accuracies with respect to different support threshold $\delta$; (4) the average of plan lengths; (5) the running time of {\imp}. We compared our {\imp} algorithm with the state-of-the-art CBP (Case Based Planning) system OAKPlan \cite{journals/aij/Serina10}. OAKPlan requires a \emph{complete} domain model and a case library as input for a new planning problem. To make OAKPlan be comparable with our {\imp} algorithm, we fed an \emph{incomplete} domain model to OAKPlan, which was the same as the input of {\imp}, instead of an complete domain model.
\subsubsection{Varying the number of plan examples}
\begin{figure}[!ht]
  \centering
  \includegraphics[width=0.45\textwidth]{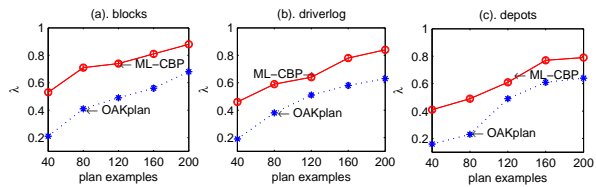}\\
  \caption{Comparison between {\imp} and OAKPlan with respect to different number of plan examples.}\label{imp_oakplans}
\end{figure}

We would like to test the change of the accuracy when the number of plan examples increasing.  We set the percentage of completeness as 60\%, and the threshold $\delta$ as 15. We varied the number of plan examples from 40 to 200 and run {\imp} to solve 100 planning problems. We calculated the accuracy $\lambda$ for each case. The result is shown in Figure \ref{imp_oakplans}.

From Figure \ref{imp_oakplans}, we found that both accuracies of {\imp} and OAKPlan generally became larger when the number of plan examples increased. This is consistent with our intuition, since there is more knowledge to be used when plan examples become larger. On the other hand, we also found that {\imp} generally had higher accuracy than OAKPlan in all the three domains. This is because {\imp} exploits the information of incomplete domain models to mine \emph{multiple} high quality plan fragments, i.e., {\imp} integrates the knowledge from both incomplete domain models and plan examples, which may help each other,  to attain the final solution. In contrast, OAKPlan first retrieves a case, and then adapts the case using the inputted incomplete domain model, which may fail to make use of valuable information from other cases (or plan fragments) when adapting the case.

By observation, we found that the accuracy of {\imp} was no less than 0.8 when the number of plan examples was more than 160.

\subsubsection{Varying the percentage of completeness}
\begin{figure}[!ht]
  \centering
  \includegraphics[width=0.46\textwidth]{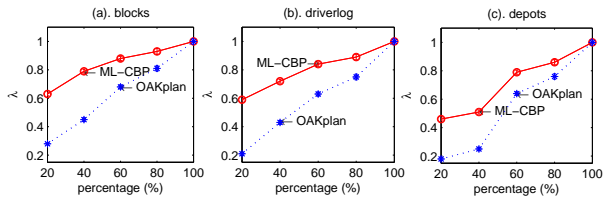}\\
  \caption{Comparison between {\imp} and OAKPlan with respect to different percentage of completeness.}\label{imp_oak_incomp}
\end{figure}

To test the change of accuracies with respect to different degrees of completeness, we varied the percentage of completeness from 20\% to 100\%, and ran {\imp} with 200 plan examples by setting $\delta=15$. We also compared the accuracy with OAKPlan. The result is shown in Figure \ref{imp_oak_incomp}.

We found both accuracies of {\imp} and OAKPlan increased when the percentage of completeness increased, due to more information provided when the percentage increasing. When the percentage is 100\%, both {\imp} and OAKPlan can solve all the solvable planning problems successfully. Similar to Figure \ref{imp_oakplans}, {\imp} functions better than OAKPlan. The reason is similar to Figure \ref{imp_oakplans}, i.e., simultaneously exploiting both knowledge from incomplete domain models and plan examples could be helpful.

By observing all three domains in Figure \ref{imp_oak_incomp}, we found that {\imp} functioned much better when the percentage was smaller. This indicates that exploiting multiple plan fragments, as {\imp} does, plays a more important role when the percentage is smaller. OAKPlan does not consider this factor, i.e., it still retrieves only one case.

\subsubsection{Average of plan length}
We calculated an average of plan length for all problems successfully solved by {\imp} when $\delta$ was 15, the percentage of completeness was 60\%, and 200 plan examples were used. As a baseline, we exploited FF to solve the same problems using the corresponding complete domain models and calculate an average of their plan length. The result is shown in Table \ref{plan_length}.
\begin{table}[!ht]
  \centering
  \caption{Average of plan length}\label{plan_length}
  \begin{tabular}{|c|c|c|c|}
    \hline
    domains & blocks & driverlog & depots \\
    \hline\hline
    {\imp} & 46.8 & 83.4 & \textbf{95.3} \\
    \hline
    FF & \textbf{35.2} & \textbf{79.2} & 96.7 \\
    \hline
  \end{tabular}
\end{table}

From Table \ref{plan_length}, we found that the plan length of {\imp} was larger than FF in some cases, such as \emph{blocks} and \emph{driverlog}. However, it was also possible that {\imp} had shorter plans than FF (e.g., \emph{depots}), since high quality plan fragments could help acquire shorter plans.

\subsubsection{Varying the support threshold}
We tested different support thresholds to see how they affected the accuracy. We set the completeness to be 60\%. The result is shown in Table \ref{threshold}. The bold parts indicate the highest accuracies. We found that the threshold could not be too high or too low, as was shown in domains \emph{blocks} and \emph{driverlog}. A high threshold may incur \emph{false negative}, i.e., ``good'' plan fragments are excluded when mining frequent plan fragments in step 3 of Algorithm \ref{algorithm:main}. In contrast, a low threshold may incur \emph{false positive}, i.e., ``bad'' plan fragments are introduced. Both of these two cases may reduce the accuracy. We can see that the best choice for the threshold could be 15 (the accuracies of $\delta=15$ and $\delta=25$ are close in \emph{depots}).
\begin{table}[!ht]
 \caption{Accuracy with respect to different thresholds.}\label{threshold}
 \centering
  \begin{tabular}{|c|c|c|c|}
  \hline
  threshold  & blocks & driverlog & depots \\
  \hline
  $\delta = 5$ & 0.80 & 0.78 & 0.73\\
  \hline
  $\delta = 15$ & \textbf{0.88} & \textbf{0.84} & 0.79 \\
  \hline
  $\delta = 25$ & 0.83 & 0.75 & \textbf{0.80} \\
  \hline
  \end{tabular}
\end{table}

\subsubsection{The running time}
\begin{figure}[!ht]
  \centering
  \includegraphics[width=0.38\textwidth]{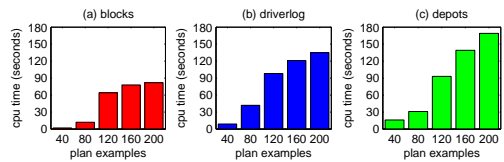}\\
  \caption{The running time of our {\imp} algorithm}\label{cputime}
\end{figure}

We show the average CPU time of our {\imp} algorithm over 100 planning problems with respect to different number of plan examples in Figure \ref{cputime}. As can be seen from the figure, the running time increases polynomially with the number of input plan traces. This can be verified by fitting the relationship between the number of plan examples and the running time to a performance curve with a polynomial of order 2 or 3. For example, the fit polynomial for \emph{blocks} is $-0.0022x^2+1.1007x-45.2000$.

\section{Conclusion}
In this paper, we presented a system called {\imp}  for doing model-lite case-based
planning. {\imp} is able to integrate knowledge from both incomplete domain models and
a library of  plan examples to produce solutions to new planning
problems. With the incomplete domain models, we first generate a
skeletal plan using of-the-shelf planners, and then mine sequential
information from plan examples to finally generate solutions. Our
experiments show that {\imp} is effective in three benchmark
domains compared to case-based planners that rely on complete domain
models.  Our approach is thus well suited for scenarios where the
planner is limited to  incomplete models of the
domain, but  does have access to a library of plans correct with
respect to the complete (but unknown) domain theory.
Our work can be seen as a contribution both to model-lite planning,
which is interested in plan synthesis under incomplete domain models,
and the original vision of case-based planning, which aimed to use a
library of cases as an extensional representation of planning
knowledge.

\newpage


\begin{thebibliography}{}

\bibitem[\protect\citeauthoryear{Alterman}{1986}]{conf/aaai/alterman86}
Alterman, R.
\newblock 1986.
\newblock An adaptive planner.
\newblock In {\em Proceedings of AAAI},  65--71.

\bibitem[\protect\citeauthoryear{Amir}{2005}]{conf/ijcai/Amir05}
Amir, E.
\newblock 2005.
\newblock Learning partially observable deterministic action models.
\newblock In {\em Proceedings of IJCAI},  1433--1439.

\bibitem[\protect\citeauthoryear{Bertoli, Pistore, and
  Traverso}{2010}]{journal/aij/Bertoli10}
Bertoli, P.; Pistore, M.; and Traverso, P.
\newblock 2010.
\newblock Automated composition of web services via planning in asynchronous
  domains.
\newblock {\em Artificial Intelligence Journal} 174(3-4):316--361.

\bibitem[\protect\citeauthoryear{Blythe, Deelman, and
  Gil}{2004}]{journal/is/Blythe04}
Blythe, J.; Deelman, E.; and Gil, Y.
\newblock 2004.
\newblock Automatically composedworkflows for grid environments.
\newblock {\em IEEE Intelligent Systems} 19(4):16--23.

\bibitem[\protect\citeauthoryear{Bryce and Weber}{2011}]{cof/icaps/bryce11}
Bryce, D., and Weber, C.
\newblock 2011.
\newblock Planning and acting in incomplete domains.
\newblock In {\em Proceedings of ICAPS}.

\bibitem[\protect\citeauthoryear{Fikes and
  Nilsson}{1971}]{journal/aij/Richard71}
Fikes, R., and Nilsson, N.~J.
\newblock 1971.
\newblock {STRIPS}: A new approach to the application of theorem proving to
  problem solving.
\newblock {\em Artificial Intelligence Journal}  189--208.

\bibitem[\protect\citeauthoryear{Hammond}{1989}]{book/Hammond89}
Hammond, K.~J.
\newblock 1989.
\newblock {\em Case-Based Planning: Viewing Planning as a Memory Task}.
\newblock San Diego, CA: Academic Press.

\bibitem[\protect\citeauthoryear{Hoffmann, Bertoli, and
  Pistore}{2007}]{conf/aaai/Hoffmann07}
Hoffmann, J.; Bertoli, P.; and Pistore, M.
\newblock 2007.
\newblock Web service composition as planning, revisited: In between background
  theoriesandinitial state uncertainty.
\newblock In {\em Proceedings of AAAI}.

\bibitem[\protect\citeauthoryear{Ihrig and
  Kambhampati}{1997}]{journal/jair/Ihrig97}
Ihrig, L.~H., and Kambhampati, S.
\newblock 1997.
\newblock Storing and indexing plan derivations through explanation-based
  analysis of retrieval failures.
\newblock {\em Journal of Artificial Intelligence Research} 7:161--198.

\bibitem[\protect\citeauthoryear{Kambhampati and Hendler}{1992}]{priar}
Kambhampati, S., and Hendler, J.~A.
\newblock 1992.
\newblock A validation-structure-based theory of plan modification and reuse.
\newblock {\em Artificial Intelligence Journal} 55:193¨C258.

\bibitem[\protect\citeauthoryear{Kambhampati}{2007}]{conf/aaai/rao07}
Kambhampati, S.
\newblock 2007.
\newblock Model-lite planning for the web age masses: The challenges of
  planning with incomplete and evolving domain theories.
\newblock In {\em Proceedings of AAAI}.

\bibitem[\protect\citeauthoryear{Nguyen, Kambhampati, and
  Do}{2010}]{conf/icaps/Nguyen10}
Nguyen, T.~A.; Kambhampati, S.; and Do, M.~B.
\newblock 2010.
\newblock Assessing and generating robust plans with partial domain models.
\newblock In {\em ICAPS Workshop on Planning under Uncertainty}.

\bibitem[\protect\citeauthoryear{Pei \bgroup et al\mbox.\egroup
  }{2004}]{journal/tkde/pei04}
Pei, J.; Han, J.; Mortazavi-Asl, B.; Wang, J.; Pinto, H.; Chen, Q.; Dayal, U.;
  and Hsu, M.-C.
\newblock 2004.
\newblock Mining sequential patterns by pattern-growth: The prefixspan
  approach.
\newblock {\em IEEE Transactions on Knowledge and Data Engineering}
  16(11):1424--1440.

\bibitem[\protect\citeauthoryear{Serina}{2010}]{journals/aij/Serina10}
Serina, I.
\newblock 2010.
\newblock Kernel functions for case-based planning.
\newblock {\em Artificial Intelligence} 174(16-17):1369--1406.

\bibitem[\protect\citeauthoryear{Sutton and Barto}{1998}]{rl-book}
Sutton, R.~S., and Barto, A.~G.
\newblock 1998.
\newblock {\em Reinforcement Learning: An Introduction}.
\newblock Cambridge, Massachusetts: MIT Press.

\bibitem[\protect\citeauthoryear{Veloso \bgroup et al\mbox.\egroup
  }{1995}]{prodigy-analogy}
Veloso, M.; Carbonell, J.; Pérez, A.; Borrajo, D.; Fink, E.; and Blythe, J.
\newblock 1995.
\newblock Integrating planning and learning: The prodigy architecture.
\newblock {\em Journal of Experimental and Theoretical Artificial Intelligence}
  7(1).

\bibitem[\protect\citeauthoryear{Yang, Wu, and
  Jiang}{2007}]{journal/aij/Yang07}
Yang, Q.; Wu, K.; and Jiang, Y.
\newblock 2007.
\newblock Learning action models from plan examples using weighted {MAX-SAT}.
\newblock {\em Artificial Intelligence Journal} 171:107--143.

\bibitem[\protect\citeauthoryear{Zaki}{2001}]{journal/ml/Zaki01}
Zaki, M.~J.
\newblock 2001.
\newblock Spade: An efficient algorithm for mining frequent sequences.
\newblock {\em machine learning} 42:31--60.

\bibitem[\protect\citeauthoryear{Zettlemoyer, Pasula, and
  Kaelbling}{2005}]{conf/aaai/Leslie05}
Zettlemoyer, L.~S.; Pasula, H.~M.; and Kaelbling, L.~P.
\newblock 2005.
\newblock Learning planning rules in noisy stochastic worlds.
\newblock In {\em Proceedings of AAAI}.

\bibitem[\protect\citeauthoryear{Zhuo \bgroup et al\mbox.\egroup
  }{2010}]{journal/aij/zhuo10}
Zhuo, H.~H.; Yang, Q.; Hu, D.~H.; and Li, L.
\newblock 2010.
\newblock Learning complex action models with quantifiers and implications.
\newblock {\em Artificial Intelligence} 174(18):1540 -- 1569.

\end{thebibliography}
\end{document}